\documentclass[final]{cvpr}

\usepackage{times}
\usepackage{epsfig}
\usepackage{graphicx}
\usepackage{amsmath}
\usepackage{amssymb}
\usepackage{float}
\usepackage{booktabs}
\usepackage{color}
\usepackage{stfloats}
\usepackage{lipsum}

\usepackage[pagebackref=true,breaklinks=true,colorlinks,bookmarks=false]{hyperref}
\graphicspath{{figures/}}
\def\cvprPaperID{****} 
\def\confYear{CVPR 2021}

\newcommand\blfootnote[1]{%
	\begingroup
	\renewcommand\thefootnote{}\footnote{#1}%
	\addtocounter{footnote}{-1}%
	\endgroup
}

\begin{document}

\title{Updatable Siamese Tracker with Two-stage One-shot Learning}

\author{XingLong Sun$^{1,2}$, GuangLiang Han$^{1*}$, Lihong Guo$^{1*}$, Tingfa Xu$^{3}$, Jianan Li$^{3}$, Peixun Liu$^{1}$\\
$^{1}$Changchun Institute of Optics, Fine Mechanics and Physics, Chinese Academy of Sciences, China\\
$^{2}$University of Chinese Academy of Sciences, China\\
$^{3}$School of Optoelectronics, Beijing Institute of Technology, China\\
{\tt\small $\{$sunxinglong,hangl,guolh$\}$@ciomp.ac.cn, $\{$ciomp$\_$xtf1,lijianan$\}$@bit.edu.cn, liupx@ciomp.ac.cn}
}

\maketitle

\blfootnote{*Corresponding author.}
\begin{abstract}
Offline Siamese networks have achieved very promising tracking performance, especially in accuracy and efficiency. However, they often fail to track an object in complex scenes due to the incapacity in online update. Traditional updaters are difficult to process the irregular variations and sampling noises of objects, so it is quite risky to adopt them to update Siamese networks. In this paper, we first present a two-stage one-shot learner, which can predict the local parameters of primary classifier with object samples from diverse stages. Then, an updatable Siamese network is proposed based on the learner (SiamTOL), which is able to complement online update by itself. Concretely, we introduce an extra inputting branch to sequentially capture the latest object features, and design a residual module to update the initial exemplar using these features. Besides, an effective multi-aspect training loss is designed for our network to avoid overfit. Extensive experimental results on several popular benchmarks including OTB100, VOT2018, VOT2019, LaSOT, UAV123 and GOT10k manifest that the proposed tracker achieves the leading performance and outperforms other state-of-the-art methods.
\end{abstract}

\section{Introduction}

Visual object tracking attracts a considerable amount of attention in the past few years, due to its wide applications in visual surveillance \cite{Conference1}, robotics \cite{Journal1}, human computer interaction \cite{Conference2} and augmented reality  \cite{Conference3}. With the state of an object in the initial frame as inference, tracking aims to predict the object’s state in each subsequent frame. Despite have realized great progress, the tracking performance still be restricted by some challenging factors, like occlusion, illumination variation, background clutters, deformation, etc.

With the development of deep learning, most of trackers have achieved very satisfactory performance, especially Siamese trackers \cite{Conference4,Conference5,Conference6,Conference7}. They formulate object tracking as learning a similarity metric between the features of the template and the candidate object patches. Based on the seminal SiamFC \cite{Conference4} and SINT \cite{Conference8}, many methods try to improve the tracking performance of Siamese trackers. Meanwhile, some \cite{Conference7,Conference9,Conference10} focus on constructing deeper backbone networks for better feature representation, while others \cite{Conference6,Conference7,Conference11} expect to design more powerful decision modules to predict the object states. Besides, a number of efforts \cite{Conference12,Conference13} study how to promote the training quality.

It is well-known that Siamese trackers are usually trained in an offline manner, which learn the template features of objects only in the first frame. In this case, they are difficult to adapt to severe appearance variations of objects, falling into tracking failure in complex scenes. To address the problem, a direct solution is to update these trackers online. However, update online it is not a simple issue due to the irregularity of appearance variations in both temporal and spatial dimensions and the inevitable noises when extracting object samples. Previous updating methods \cite{Conference5,Conference14} often struggle with challenging update requirements, and thus make trackers drift away \cite{Conference15}. Recently, a few works such as UpdateNet \cite{Conference16} utilize Deep Neural Networks (DNNs) to update online, which are more effective in lifting tracking performance with the advantage of adaptive feature representation. Whereas, an essential problem is that updating networks and trackers are enough structurally separated, which cannot benefit from joint training and cooperate each other in the best way.

Instead of taking advantage of extra update modules, can we present a specific Siamese network which is capable of completing high-quality online update by itself? Our work finds that the answer is affirmable by exploring one-shot learning \cite{Conference17,Conference18}, which devotes itself to learning object attributes with as few samples as possible. According to the description in \cite{Conference6}, Siamese network is an implicit expression of the one-shot learner \cite{Conference17}. Inspired by this opinion, this paper first proposes a novel two-stage one-shot learner, which is able to learn object information with samples from diverse stages. Then, a novel Siamese network is designed with the guideline of the learner, which would automatically update the exemplar when tracking an object. Specifically, in addition to learn initial features with the template branch, we adopt an extra branch to capture the object features in subsequent frames. The features from diverse phases are combined by a residual module to generate a reliable tracking template. Moreover, we also present a multi-aspect loss function to efficiently train the network in an end-to-end manner. At last, the proposed Siamese tracker gets outstanding performance by taking tracking results as updating samples.

The main contributions of this paper are summarized as follows:

A novel two-stage one-shot learner is proposed for visual tracking, which has a capacity to continually learn sample attributes in diverse stages to adjust the network.

We present an updatable Siamese tracker based on the two-stage one-shot learner. By fusing multi-frame object features via residual learning, the tracker is able to track the current object with a more suitable template.

Numerous experimental results on some popular testing datasets demonstrate that the proposed tracker can realize state-of-the-art performance. 

\section{Related works}
In this section, we briefly review the recent literatures of Siamese trackers, as well as discuss the related topics on online update and one-shot learning.
\subsection{Siamese Trackers}
As a popular tracking paradigm, Siamese networks have been discussed thoroughly in previous researches. A pioneering work is SiamFC \cite{Conference4}, which matched the exemplar and the candidate sample patches with a correlation layer. After that, some literatures tried to improve training quality by using adversarial learning \cite{Conference12} or distractor-aware sampling \cite{Conference13}. Besides, a notable direction is to design powerful decision modules. The most successful case is Region Proposal network (RPN) \cite{Conference21,Conference6,Conference22,Conference23}, which could classify the objects from backgrounds and regress their bounding boxes simultaneously. Other effective decision modules were also exploited for Siamese trackers, i.e., anchor-free models \cite{Conference11,Conference24}, segmentation networks \cite{Conference10}, corner detection modules \cite{Conference43}, etc. Another issue is to increase the depth of backbone to improve feature representation, and the key is to break the limitation of padding. SiamPRN++ \cite{Conference7} adopted ResNet \cite{Conference25} as backbone by introduce spatial aware sampling, while SiamDW \cite{Conference9} designed a novel residual unit without padding operation. Despite achieving promising results, all the above approaches have no ability to finish adaptive online update, which is very fatal to track an object in complex scenes.
\subsection{Online update methods}
Online update plays a very vital role to maintain tracking performance, especially for offline deep trackers \cite{Conference7,Conference11,Conference26}. As a result, a great number of approaches are presented to complete the issue. Deep discriminated trackers \cite{Conference26,Conference27,Journal2} were usually updated by fine-tuning model online, which are extremely time-consuming and easy to overfit due to limited training samples, even if reinforcement learning \cite{Conference28} or adversarial learning \cite{Conference29} has been employed. For Siamese trackers, some methods updated the initial exemplar through running moving \cite{Conference5} or feature fast transformation [15]. Whereas, they often failed to satisfy complex updating demands with the impact of irregular appearance variations and sample noises. Nowadays, neural networks gradually became an alternative answer to update deep trackers. Among these, UpdateNet \cite{Conference16} designed a powerful deep updating module which could benefit from the training on large-scale video datasets, and Meta-updater \cite{Conference15} attempted to update long-term trackers with the LSTM \cite{Journal3} based on meta learning. Deep networks are more adaptive and reliable for update, but they are still difficult to update Siamese trackers in an optimal way since they are completely separated with tracking models.  

\subsection{One-shot learning}

One-shot learning focuses on learning a generic function on various issues, such that it can be generalized to a novel issue with only a very few of samples. Various schemes were employed to finish the task, such as memory networks \cite{Conference30,Journal4}, metric learning \cite{Conference31,Conference32}, optimization methods \cite{Conference18,Conference33} and parameter learning network \cite{Conference17}, which have been applied widely in the field of visual tracking. For example, GradNet \cite{Conference20} presented a gradient-guided network to update the exemplar for Siamese network, while \cite{Conference15} proposed a meta-updater based on a memory network for long-term tracking. Besides, Meta-Tracker \cite{Conference19} accelerated the process of online domain training by exploring model-agnostic meta-learning (MAML) \cite{Conference33}, which was also used to fine-tune the model weights in MetaRTT \cite{Journal5}. Different from these methods, our work is most related to \cite{Conference17}, which constructed the second network as a learnet to predict the parameters of the primary network. By reformulating Siamese network with the learner, this paper develops a novel two-shot one-shot learner for updating Siamese tracker.

\begin{figure}[t]
	\begin{center}
		\includegraphics[width=0.8\linewidth]{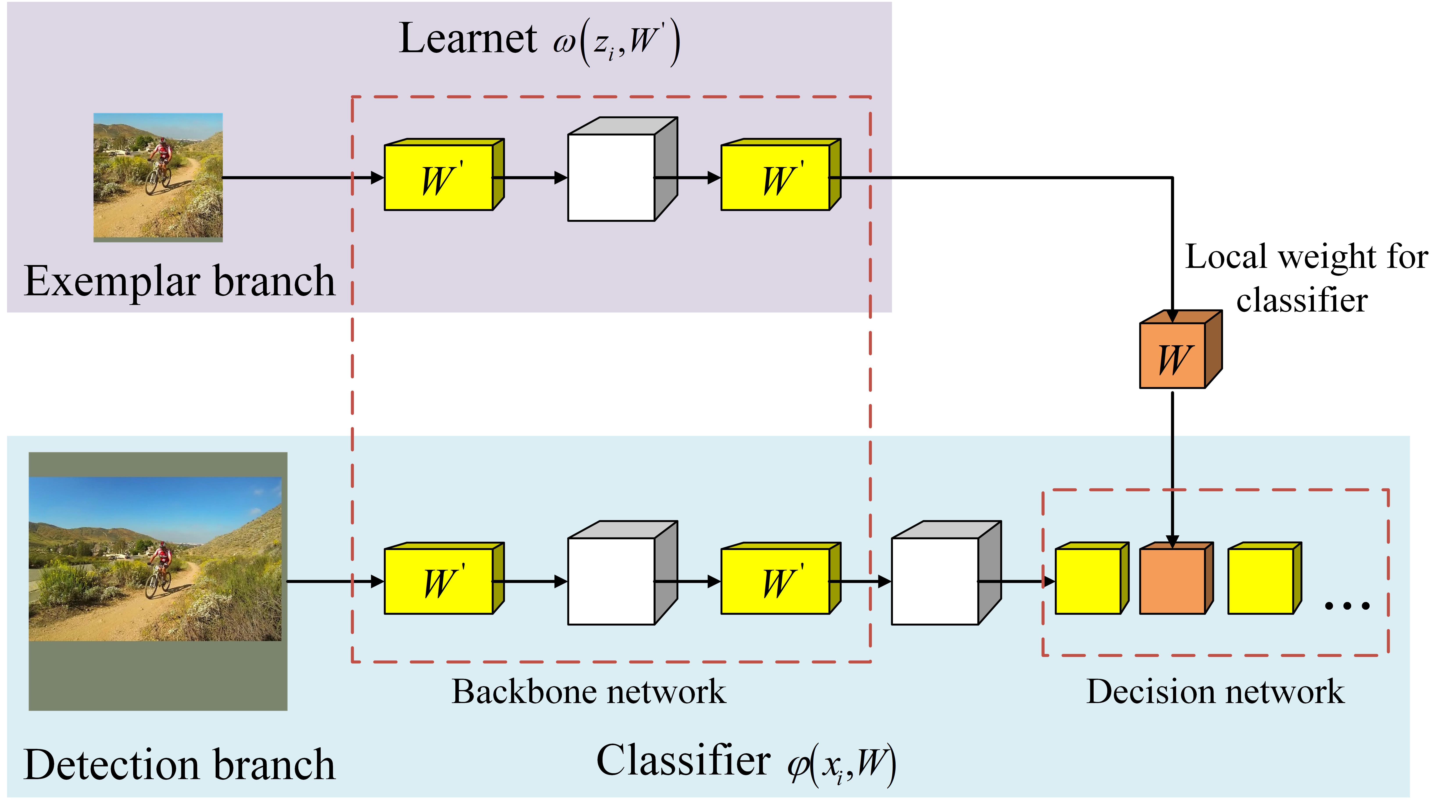}
	\end{center}
	\caption{one-shot learning formulation of Siamese network. The exemplar branch is viewed as a learnet, while the detection branch is a classifier for identifying object from background.}
	\label{fig:one-shot}
	\label{fig:onecol}
\end{figure}
\section{Two-stage One-shot Learning}
In addition to the similarity measuring function, Siamese network is also interpreted as an one-shot learner in some previous works \cite{Conference6}. In this section, we first formulate Siamese network with one-shot learning, and then propose a two-stage one-shot learner for designing updatable Siamese network.

\subsection{One-shot learning formulation}
In field of deep learning, a representative framework for object tracking is the discriminated classifier $\varphi(x, W)$ \cite{Conference26,Conference27}. Its learning object is to find the parameter $W$ that can minimize the total loss $\mathcal{L}$ over the training dataset
\begin{equation}
\min _{W} \frac{1}{n} \sum_{i=1}^{n} \mathcal{L}\left(\varphi\left(x_{i}, W\right), \ell_{i}\right) \label{eqt1}
\end{equation}
in which, $n$ denotes the number of training samples in the dataset, and $\ell_{i}$ is the class label of sample $x_{i}$.  The classifier is pretty competitive in identifying the tracking object, but it always requires online training to learn object appearance information. The computational burden is fatal to real-time tracking requirement.

To break the limitation, another tracking framework, as well Siamese network has been proposed \cite{Conference4,Conference7}, whose learning object is to learn a similarity matching function between exemplar patch and candidate region patches
\begin{equation}
\min _{W} \frac{1}{n} \sum_{i=1}^{n} \mathcal{L}\left(\zeta\left(\varphi^{\prime}\left(x_{i}, W^{\prime}\right), \varphi^{\prime}\left(z_{i}, W^{\prime}\right)\right), \ell_{i}\right)  \label{eqt2}
\end{equation}
in which, $z_{i}$ denotes the exemplar patch, and $x_{i}$ represents the search region patch. Besides, function $\varphi^{\prime}$ and $\zeta$ denote the feature subnetwork and the matching decision subnetwork, respectively. By training with large-scale video sequences, Siamese networks can quickly recognize the most similar candidate patch with the initial exemplar.

Actually, Siamese model can be reinterpreted as a one-shot learning classifier. For a classifier $\varphi(x, W)$, if its parameter $W$ can be learned with a single sample $z_{i}$ from the class of interest, which can be translated in a one-shot learner, i.e., learning to learn. To incorporate sufficient prior knowledge, a meta-learning function $\omega$ is proposed to map $\left(z, W^{\prime}\right)$ to $W$ \cite{Conference17}.
\begin{equation}
\min _{W \prime} \frac{1}{n} \sum_{i=1}^{n} \mathcal{L}\left(\varphi\left(x_{i}, \omega\left(z_{i}, W^{\prime}\right)\right), \ell_{i}\right) \label{eqt3}
\end{equation}
By training on millions of small one-shot learning issues offline, the learnet $\omega$ is easy to predict the parameter $W$ by non-iterative feed-forward propagation. For Siamese model, if we regard the detection branch as a classifier, the exemplar branch $\varphi^{\prime}\left(z_{i}, W^{\prime}\right)$ could be deemed to be a meta-learner $\omega$ to compute the local parameter $W$ of the classifier. In this view, Siamese network is an explicit realization of one-shot learning, though the meta-learner $\omega$ is shared for extracting the features of search region, as displayed in Figure \ref{fig:one-shot}.

\subsection{Two-stage One-shot learner}
It is convenient for Siamese network to capture the appearance information of object through utilizing its initial template, which is adopted to directly estimate the object state during tracking. However, the object appearance may undergo severe and unknown variations, and thus a tracker must continually learn the attribute information of object in subsequent frames. It is impossible to the one-shot learner of Siamese network, since it only can perform one-shot learning in one stage. As a result, Siamese network generally struggles with online update problem. 

In this work, we continue to exploit the one-shot learning architecture to solve online update problem. We believe there is a one-shot learner that is able to learn object appearance information in different stages, just like the one-shot learner of Siamese network learns the information in the initial frame. Based on the inspiration, we design a novel one-shot learner for tracking, named two-stage one-shot learner (TOL). The learner can combine the samples with diverse attributes for predicting the parameter $W$ of classifier. Its learning object can be formulated as
\begin{equation}
\min _{W^{\prime}} \frac{1}{n} \sum_{i=1}^{n} \mathcal{L}\left(\varphi\left(x_{i}, \omega\left(z_{i}, u_{i}, W^{\prime}\right)\right), \ell_{i}\right) \label{eqt4}
\end{equation}
where, $z_{i}$ and $u_{i}$ indicates the one-shot learning samples from multiple stages, respectively. When tracking an object, we can take the initial object exemplar as the sample $z_{i}$, and regard the tracking results in the subsequent frames as the sample $z_{i}$. In this way, Siamese tracker is able to capture object information sequentially during the whole tracking process, and achieve more stable and robust online tracking.
\begin{figure*}
	\begin{center}
		\includegraphics[width=0.75\linewidth]{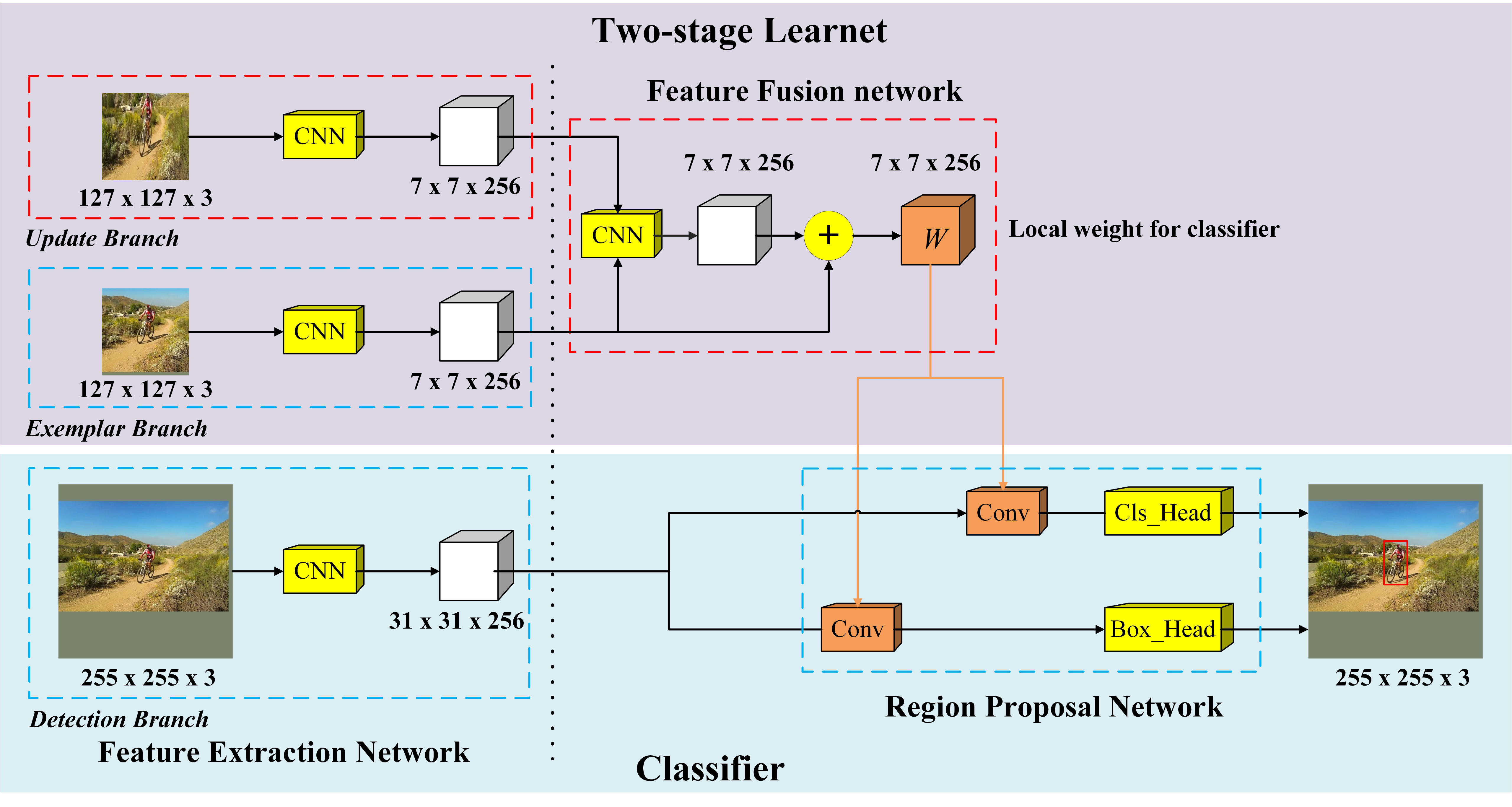}
	\end{center}
	\caption{Overview of the proposed framework, consisting of the shared backbone network, feature fusion network and region proposal network. The backbone networks in both exemplar branch and update branch as well as the feature fusion network works as a two-stage learnet, which has an ability to generate the part weights of the RPN block in classifier (detection branch).}
	\label{fig:framework}
\end{figure*}

\section{Updatable Siamese Network}
In this section, we depict the proposed updatable Siamese network, which is built upon the two-stage one-shot learner. It mainly consists of the network construction, the training method and the strategy of online update.

\subsection{Framework}
The framework of our presented network is illustrated in Figure \ref{fig:framework}. In the scheme, the backbone is responsible for simultaneously extracting features from exemplar, update sample and search region samples, and the feature fusion network combines the features of the exemplar and update sample by residual learning, which would output a more appropriate template for tracking the current object. At last, the RPN blocks are applied to predict the final state of object.

\textbf{Backbone network.} Various deep neural networks have been successfully applied in Siamese trackers \cite{Conference6,Conference7}. In our tracker, we adopt ResNet-50 \cite{Conference25} as the backbone network, and employ some modifications to it. To ensure tracking efficiency, the last residual block in standard ResNet-50 is first removed, and only the first three residual blocks are reserved for feature extraction. Then, the sampling stride in the third block is adjusted from 16 to 8, while the dilated convolution is utilized to enlarge the receptive field. These adjustments are very necessary since tracking need higher feature resolution for accurate location.
 
The network layers in different depths vary in receptive fields and abstracting levels, which can provide complementary features for visual tracking. The shallower layer can more spatial detailed features for location, while the deeper layer contributes to detect the object by encoding high-level semantic patterns. To exploit the characteristic, we take advantage of multi-layer features from the last two blocks, and add a 1x1 convolutional layer to each of block end to compress the channel number to 256. In exemplar and update branches, we only make use of the features on center 7×7 regions.

\textbf{Region proposal network}. Region proposal network is a popular detection module \cite{Conference21}, which has been extensively discussed in most of previous trackers \cite{Conference7,Conference22}. There are two diverse task branches in it: one for object classification, and the other for object location regression. In each branch, two input feature maps are first compared with a depth-wise correlation layer. In view of one-shot learning, the operation is a very critical step for adapting to the new tracking object, where the exemplar features are adopted as the weights of a convolutional layer  in primary classifer to detect the candidate patches. Then, a classification and a regression heads are introduced into corresponding branches to output tracking results. To benefit from multi-layer features, we introduce two RPN blocks for the last two blocks of backbone, whose outputted maps are aggregated with learnable weights. 

\textbf{Feature fusion network.} After extracting the features of initial template and update sample, we still need a module to combine these features. In consequence, we present a feature fusion network, which is constructed by cascading three 1×1 convolutional layers with different channels of 256, 128 and 256. Each of the first two convolutional layers is followed by a ReLU. The update sample features can provide the latest object appearance information, while the exemplar covers the initial object ground-truth information, which is a vital guideline for the network to learn how to take advantage of the latest object appearance. As a result, the features of exemplar and update sample are first concatenated, and then are sent into the network. We believe that the best updating way is to fine-tune the initial exemplar with the latest object features, since the initial exemplar is the only non-polluting sample for object. Hence, a residual learning structure is explored, where the fusing outputs are accumulated with the initial template features. 

\textbf{Relation to two-stage one-shot learning.} The updatable Siamese network is explicit representation of the proposed two-stage one-shot learner. The group of exemplar branch, update branch and feature fusion network plays the role of the two-stage learnet $\omega\left(z_{i}, u_{i}, W^{\prime}\right)$ in Eq. \ref{eqt4}, while the detection branch and RPN blocks can be regarded as a primary classifier, as depicted in Figure \ref{fig:framework}. The learnet is implement based on residual learning
\begin{equation}
\omega\left(z_{i}, u_{i}, W^{\prime}\right)=zf+M\left(zf, uf, W_{2}^{\prime}\right)   \label{eqt5}
\end{equation}
in which, $zf$ and $uf$ denotes the deep features of exemplar $z_{i}$ and update sample $u_{i}$ output from backbone $\varphi^{\prime}\left(*, W_{1}^{\prime}\right)$. $M$ represents the feature fusion network with weights of $W_{2}^{\prime}$. By combining object appearance information from diverse stages, the learnet has an ability to predict the local weights of the classifier more efficiently. Siamese network learns to match candidate search patches with the output exemplar of two-stage lernnet, rather than the initial object template
\begin{equation}
\min _{W} \frac{1}{n} \sum_{i=1}^{n} \mathcal{L}\left(\zeta\left(\varphi^{\prime}\left(x_{i}, W_{1}^{\prime}\right), \omega\left(z_{i}, u_{i}, W^{\prime}\right)\right), \ell_{i}\right)   \label{eqt6}
\end{equation}
in this way, the propose Siamese tracker would possess an update ability to deal with complex tracking scenes.
\begin{figure}[t]
	\begin{center}
		\includegraphics[width=1.0\linewidth]{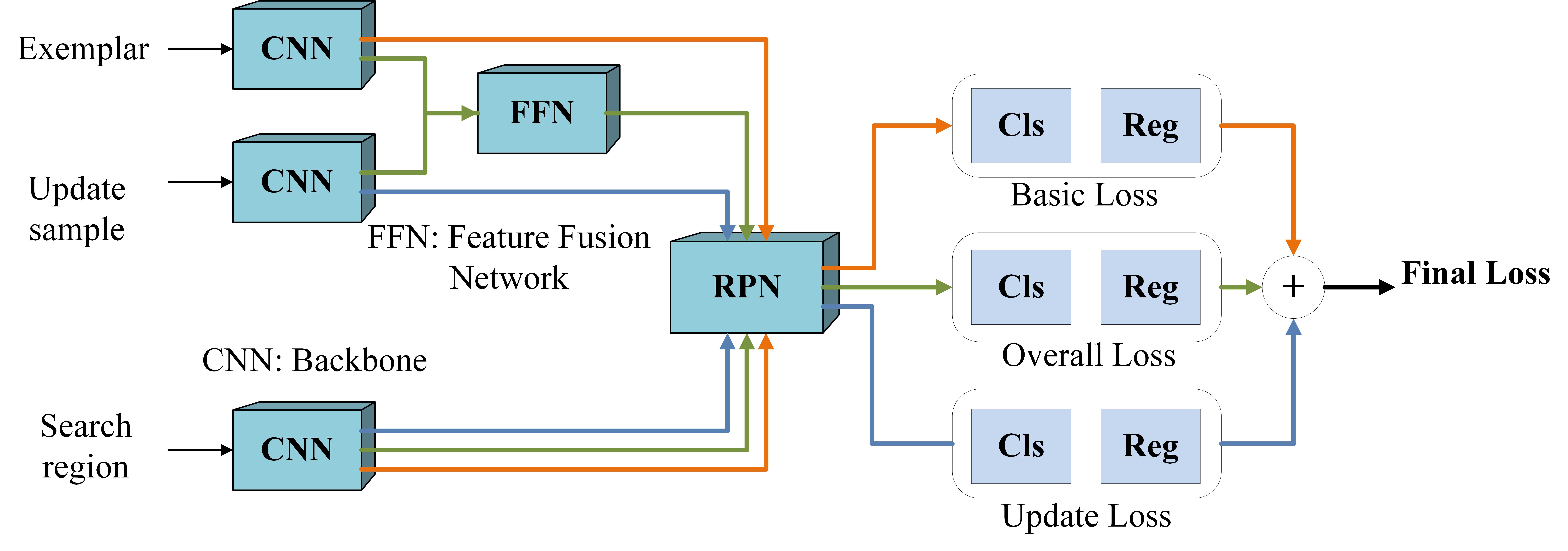}
	\end{center}
	\caption{Flowchart of the proposed Multi-aspect loss training.}
	\label{fig:train}
\end{figure}
\subsection{Multi-aspect loss training}  \label{losses}
Traditional Siamese networks are generally trained via extracting exemplar and search region samples from the same video sequences. But for the proposed network, we require to crop novel update samples from images between exemplar frames and detection frames. we also augment the update samples with some transformations such as rotation, blurring, occlusion, etc., which is valuable for our network to learn to resist sample noise.

We observe that it is very easy to overfit if we train the proposed model like training traditional Siamese networks. The main reason is that our network is composed by two units: a basic Siamese tracker, and an updater for adjusting the tracker online. If we train the network only with an overall loss, the learning may be in a dilemma to balance the function of two units. We propose a multi-aspect loss function to improve training quality, which is displayed in Figure \ref{fig:train}. 

In the first aspect, a basic loss function $\mathcal{L}_{\text {basic}}$ is designed by directly matching the exemplar and the search region samples, which will help the network to possess the basic tracking ability. With the loss, our tracker is able to work without online update like previous Siamese trackers. In the second aspect, we construct a loss $\mathcal{L}_{\text {update}}$ directly based on the update and the search region samples. It is because the update samples can also be considered as the object template, which can provide complementary training data with the exemplar samples due to the sampling gap difference. In the last aspect, we define an overall loss $\mathcal{L}_{\text {overall}}$ to learn how to predict the object state with the exemplars outputted by feature fusion network. The loss guide updater to adjust the basic tracker, which is used to train the whole network.

The above losses of $\mathcal{L}_{\text {basic}}$, $\mathcal{L}_{\text {update}}$ and  $\mathcal{L}_{\text {overall}}$ are computed with
\begin{equation}
\mathcal{L}_{p t}=\mathcal{L}_{c l s}\left(C_{p t}, \ell_{c}\right)+\lambda \mathcal{L}_{r e g}\left(L_{p t}, \ell_{L}\right) \label{eqt7}
\end{equation}
where,  $C_{p t}$ and $C_{p t}$ are the aggregated classification and regression response maps of PRN blocks corresponding to the aspect of $\text {pt}\in\{\text {basic, update, overall}\}$. $\mathcal{L}_{cls}$  is the Cross Entropy loss and $\mathcal{L}_{reg}$ is the Smooth $L1$ loss. $\ell_{c}$ and $\ell_{L}$ represent the classification and the regression ground-truths, respectively. $\lambda$ denotes a weight factor for balancing two kinds of losses. At last, the final loss function can be defined as the sum of the above losses
\begin{equation}
\mathcal{L}_{f}=\mathcal{L}_{\text {basic }}+\mathcal{L}_{\text {update }}+\mathcal{L}_{\text {overall }}  \label{eqt8}
\end{equation}

By adopting the multi-aspect loss function, we can get an outstanding Siamese tracker with powerful online update ability.

\begin{table*}	
	\begin{center}
		\setlength{\tabcolsep}{1.5mm}{		
		\begin{tabular}{l|cccccccccc}
			\toprule[1.5pt] 
			Method & UPDT & SiamPRN & ATOM & Siam-RCNN & SiamRPN++ & DiMP & PG-Net & CGACD & SiamBAN & \textbf{Ours} \\
			\midrule[1pt] 
			EAO$\uparrow$ & 0.379 & 0.383 & 0.401 & 0.408 & 0.414 & 0.440 & 0.447 & \textcolor[rgb]{0,1,0}{0.449} & \textcolor[rgb]{0,0,1}{0.452} & \textcolor[rgb]{1,0,0}{0.459}\\
			Accuracy$\uparrow$ & 0.536 & 0.588 & 0.590 & \textcolor[rgb]{0,1,0}{0.609} & 0.600 & 0.597 & \textcolor[rgb]{1,0,0}{0.618} & \textcolor[rgb]{0,0,1}{0.615} & 0.597 & 0.604 \\
			Robustness$\downarrow$ & 0.184 & 0.276 & 0.201 & 0.220 & 0.234 & \textcolor[rgb]{1,0,0}{0.152} & 0.192 & \textcolor[rgb]{0,1,0}{0.173} & 0.178 & \textcolor[rgb]{0,0,1}{0.164}\\
			\bottomrule[1.5pt]
		\end{tabular}}
	\end{center}
	\caption{Comparison with some state-of-the-art trackers on VOT2018. The best three results are highlighted in \textcolor[rgb]{1,0,0}{red}, \textcolor[rgb]{0,0,1}{blue} and \textcolor[rgb]{0,1,0}{green} fonts.}	\label{tab:vot2018}
\end{table*}

\begin{table*}[ht]	
	\begin{center}
		\setlength{\tabcolsep}{1.2mm}{		
			\begin{tabular}{l|cccccccccc}
				\toprule[1.5pt] 
				Method & MemDTC & SiamMASK & SiamRPN++ & SiamDW-ST & ATOM & CLNet & 	DCFST & DIMP & SiamBAN & \textbf{Ours} \\
				\midrule[1pt] 
				EAO$\uparrow$ & 0.228 & 0.287 & 0.287 & 0.299 & 0.301 & 0.313 & 0.317 & \textcolor[rgb]{0,1,0}{0.321} & \textcolor[rgb]{0,0,1}{0.327} & \textcolor[rgb]{1,0,0}{0.329}\\
				Accuracy$\uparrow$ & 0.485 & 0.594 & 0.599 & 0.600 & \textcolor[rgb]{0,0,1}{0.603} & \textcolor[rgb]{1,0,0}{0.606} & 0.585 & 0.582 & \textcolor[rgb]{0,1,0}{0.602} & 0.597 \\
				Robustness$\downarrow$ & 0.587 & 0.461 & 0.482 & 0.467 & 0.411 & 0.461 & \textcolor[rgb]{0,0,1}{0.376} &\textcolor[rgb]{1,0,0}{0.371} & 0.396 & \textcolor[rgb]{0,1,0}{0.381}\\
				\bottomrule[1.5pt]
		\end{tabular}}
	\end{center}
	\caption{Comparison with some state-of-the-art trackers on VOT2019. The best three results are highlighted in \textcolor[rgb]{1,0,0}{red}, \textcolor[rgb]{0,0,1}{blue} and \textcolor[rgb]{0,1,0}{green} fonts.} \label{tab:vot2019}
\end{table*}

\subsection{Online update} \label{sect:update}
Our updatable Siamese tracker is easily updated with the tracking results. But we require to select real reliable update samples to suppress the noise pollution, even if our tracker can tolerate update noise to some extent. Hence, we save the latest tracking result with confidence more than $T_m$ as the update sample. Then, we update tracker every $N$ frames to introduce object-special information in time.

\section{Experiments}
\subsection{Implementation Details}
Our work is implemented using PyTorch on a PC with one NVIDIA Titan XP GPU and a Xeon E5-2690 2.60GHz CPU.

\textbf{Training.} The proposed SiamTOL tracker is trained on the datasets of ImageNet VID \cite{Journal6}, YouTube-BB \cite{Conference34}, ImageNet DET \cite{Journal6}, COCO \cite{Conference35}, LaSOT \cite{Conference36} and GOT10k \cite{Journal7}. The sizes of exemplar and update samples are both set to $127 \times 127$, while the size of search region is $255 \times 255$. The training sample pairs are labelled with the protocol described in \cite{Conference13}. Our backbone is initialized with the weights pre-trained on ImageNet, and the first two layers are always frozen during training. We train the network using the stochastic gradient descent (SGD) optimizer with a momentum of 0.9 and a weight decay of 0.0005. The network is trained 20 epochs with a minibatch of 32, each of which includes 600k sample pairs. We use a warm-up training strategy with the learning rate varying from 0.001 to 0.005 in the first 5 epochs, and then decaying from 0.005 to 0.0005 in the last 15 epochs. Besides, the backbone is only trained in the last 10 epochs, and whose learning rate is 16 times smaller than other modules. The hyperparameter $\lambda$ of losses in Eq. \ref{eqt7} are set to 1.2.

\textbf{Testing.} In the inference, once we extract the exemplar features in the initial frame with the backbone, the object tracking can be performed automatically by continuously matching it to the features of search regions, which are cropped based on the object state of previous frame. Online update is executed following the Sect. \ref{sect:update}. Meanwhile, the sampling threshold of $T_m$ is set to $0.9$. The span of periodic update is set to $10$ frames. Besides, to smooth the movements of object, we also adopt cosine window penalty and scale change penalty to adjust the predicted boxes.

\begin{figure}[t]
	\begin{center}
		\includegraphics[width=1.0\linewidth]{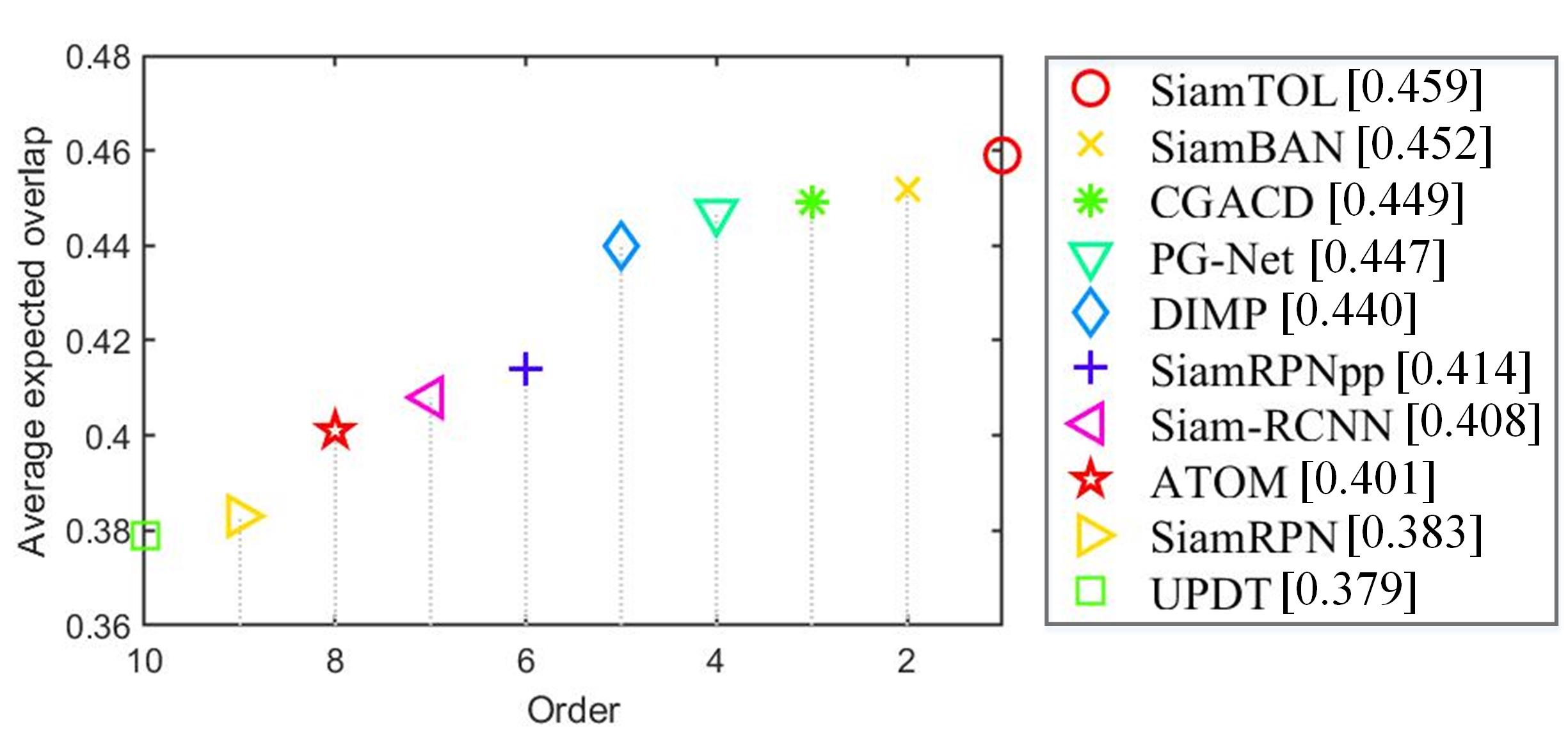}
	\end{center}
	\caption{Expected averaged overlap performance on VOT2018.}
	\label{fig:vot2018eao}
\end{figure}
\subsection{Comparison with the state-of-the-arts}
To display the superiority of our SiamTOL tracker, we compared it with the state-of-the-art trackers on six popular datasets.

\textbf{VOT2018 Dataset \cite{Conference37}.} We evaluated the SiamTOL tracker on popular Visual Object Tracking challenge 2018 [12], which consists of 60 sequences covering various challenging cases. 
Referring to the standard evaluation protocol of VOT2018, the performance of all trackers is measured by Accuracy, Robustness and the Expected Average Overlap (EAO), which is overall evaluation of accuracy and robustness. The comparison results with some state-of-the-art trackers are reported on Table \ref{tab:vot2018} and Figure \ref{fig:vot2018eao}. We observe that our updatable Siamese tracker achieve the best performance in term of EAO. Especially, SiamBAN \cite{Conference11} produce the top-ranked EAO score among previous trackers by exploiting more powerful anchor-free model for Siamese network, but our method still can slightly outperform it on overall performance with the advantage of online update. Besides, PG-Net obtains the most satisfactory accuracy, but the failure rate of our tracker decreases by 2.8$\%$ compared to the algorithm.

\begin{table*}	
	\begin{center}
		\setlength{\tabcolsep}{2mm}{		
			\begin{tabular}{l|cccccccccc}
				\toprule[1.5pt] 
				Method & MDNet & ECO & C-COT & SiamFC & SPM & SiamRPN++ & Siammask & SiamCAR & \textbf{Ours} \\
				\midrule[1pt] 
				AO$\uparrow$ & 0.299 & 0.316 & 0.325 & 0.374 & 0.513 & \textcolor[rgb]{0,1,0}{0.517} & 0.514 & \textcolor[rgb]{1,0,0}{0.569} & \textcolor[rgb]{0,0,1}{0.526} \\
				$SR_{0.5}$   & 0.303 & 0.309 & 0.328 & 0.404 & 0.593 & \textcolor[rgb]{0,1,0}{0.616} & 0.587 & \textcolor[rgb]{1,0,0}{0.670} & \textcolor[rgb]{0,0,1}{0.628} \\
				$SR_{0.75}$  & 0.099 & 0.111 & 0.107 & 0.144 & \textcolor[rgb]{0,1,0}{0.359} & 0.325 & \textcolor[rgb]{0,0,1}{0.366} & \textcolor[rgb]{1,0,0}{0.415} & 0.334 \\
				\midrule[1pt] 
				FPS & 1.5 & 2.6 & 0.7 & 25.8 & \textcolor[rgb]{0,0,1}{72.3} & 49.8 & \textcolor[rgb]{0,1,0}{61.3} & 52.2 & \textcolor[rgb]{1,0,0}{76.6} \\
				\bottomrule[1.5pt]
		\end{tabular}}
	\end{center}
	\caption{ comparison with state-of-the-art trackers on GOT-10k. The best three results are highlighted in \textcolor[rgb]{1,0,0}{red}, \textcolor[rgb]{0,0,1}{blue} and \textcolor[rgb]{0,1,0}{green} fonts.}	\label{tab:got10k}
\end{table*}

\begin{figure}[t]
	\begin{center}
		\includegraphics[width=1.0\linewidth]{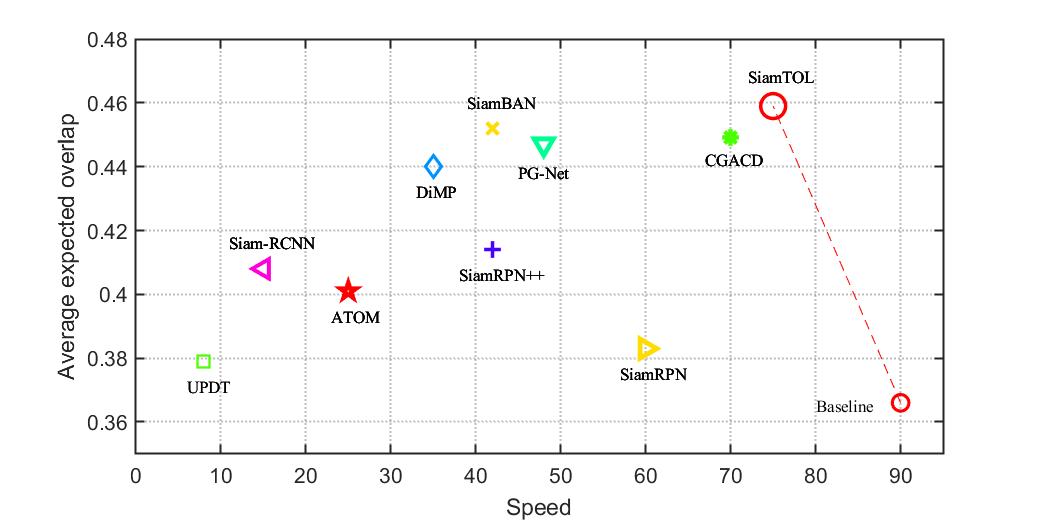}
	\end{center}
	\caption{The comparison of the quality and the speed of some state-of-the-art trackers on VOT2018. The baseline is the simplified SiamRPN++ composed by our utilized backbone and RPN networks}
	\label{fig:vot2018eao-speed}
\end{figure}

\begin{figure}[t]
	\begin{center}
		\includegraphics[width=0.49\linewidth]{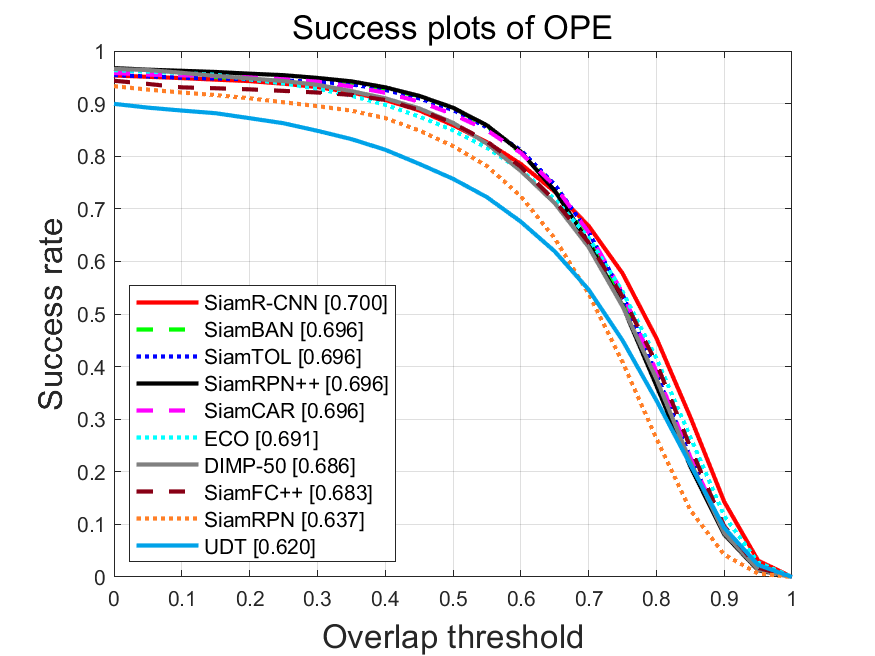}
		\includegraphics[width=0.49\linewidth]{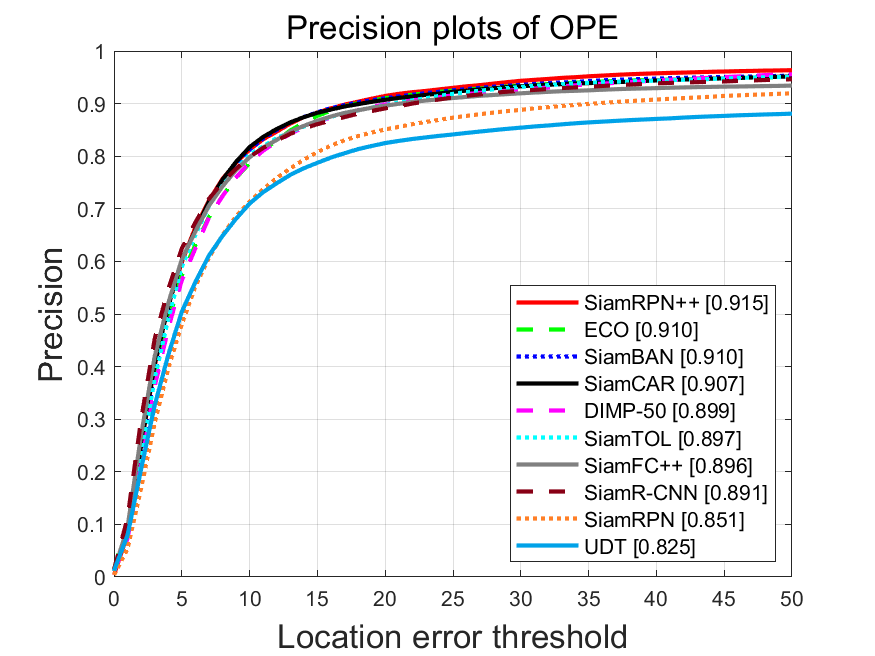}
	\end{center}
	\caption{Success plots and precision plots on OTB2015 dataset}
	\label{fig:auc_otb100}
\end{figure}
The proposed approach takes simplified SiamRPN++ as the baseline. Compared to the standard SiamRPN++ \cite{Conference7}, our method receives 4.5$\%$ improvements on EAO. The increasement mainly sources from the online update ability of our SiamTOL tracker, which provides 7$\%$ gains on robustness. This proves that the presented two-stage one-shot learning is very effective in updating Siamese network.

\textbf{Quality vs Speed.} As displayed in Figure \ref{fig:vot2018eao-speed}, we describe the EAO score with respect to the Frames-Per-Second (FPS) on VOT2018, in which our tracker is measured on TITAN X GPU. According to the figure, our tracker is superior to all comparison methods in tracking quality with a highest speed of 75 FPS. Referred to the baseline, the proposed approach obtains 9.3$\%$ improvements on EAO score, but the speed is only reduced by 15 FPS. 

\textbf{VOT2019 Dataset \cite{Conference38}.} The dataset is an updated version of VOT2018 dataset, which replaced 20 $\%$ original sequences with more representative videos. We presented the results of all trackers on Table \ref{tab:vot2019}. Among all previous methods, CLNet \cite{Conference41} achieves the best accuracy, and DiMP \cite{Conference42} is superior to others on robustness. However, CLNet is powerless in robustness, and our method realizes substantial gains of 10$\%$ on robustness over it.  For DiMP, although its robustness is slight than ours, but our tracker is more outstanding in EAO and accuracy. Compared to the SiamRPN++ \cite{Conference7}, the SiamTOL Tracker outperforms it by 10$\%$ on robustness and 4.5$\%$ on EAO. This shows that our updatable tracker can perform better than the other state-of-the-art methods.

\textbf{OTB-2015 Dataset \cite{Journal8}.} The dataset is one of the most widely used benchmark for object tracking including 100 video sequences. It covers 11 kinds of challenging factors, such as fast motion, motion blur, occlusion, and so on. In the dataset, the Area Under Curve (AUC) of success plots and the Precision score is used to evaluate tracking performance. The performance of all trackers is displayed in Figure \ref{fig:auc_otb100}. Benefit from online update, the proposed tracker is superior to most of compared methods and achieves similar results with SiamRPN++ and SiamBAN, although a more compact backbone is employed.  

\begin{figure}[t]
	\begin{center}
		\includegraphics[width=0.49\linewidth]{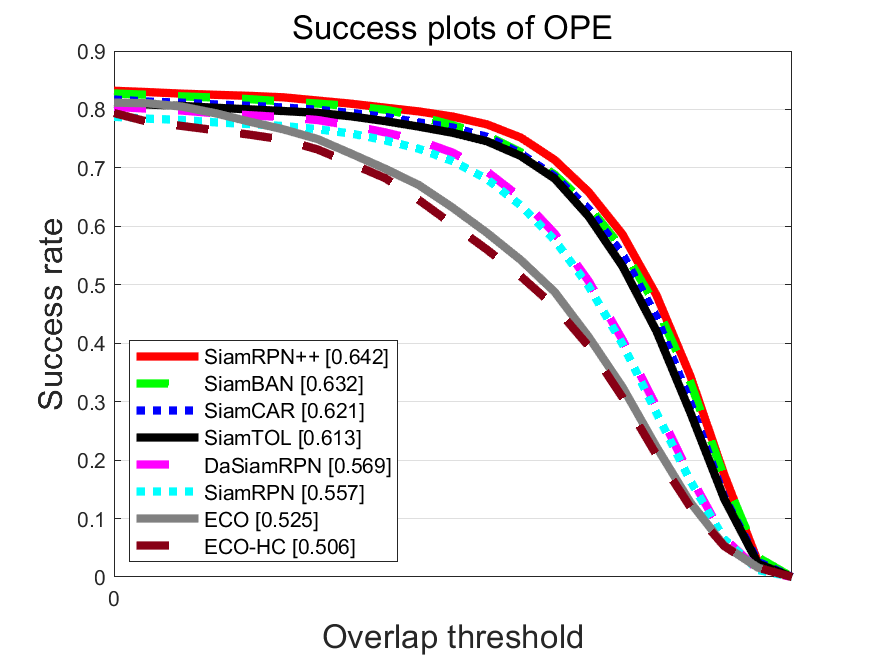}
		\includegraphics[width=0.49\linewidth]{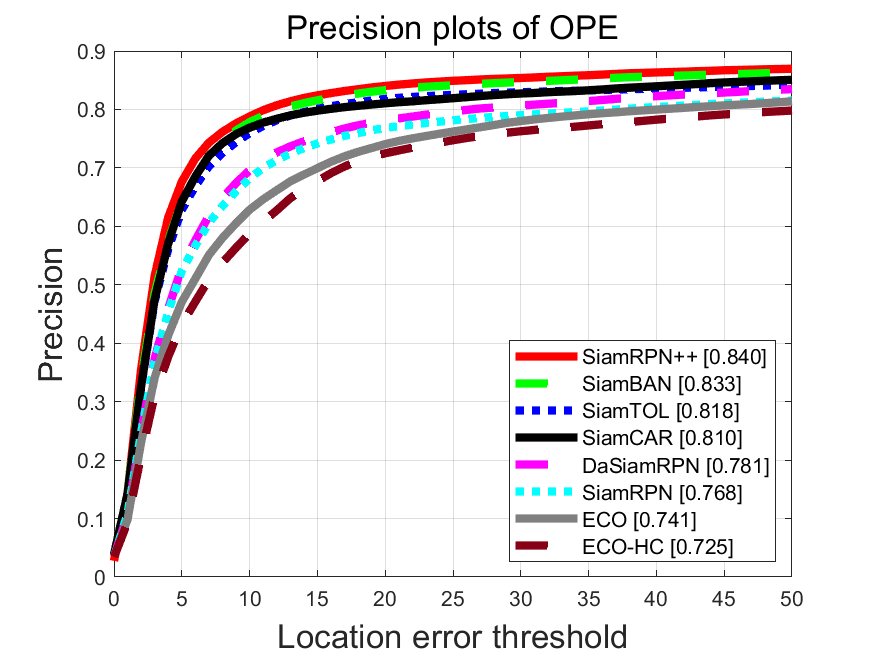}
	\end{center}
	\caption{Success and precision plots on UAV123 dataset}
	\label{fig:auc_UAV123}
\end{figure}
\begin{figure}[t]
	\begin{center}
		\includegraphics[width=0.49\linewidth]{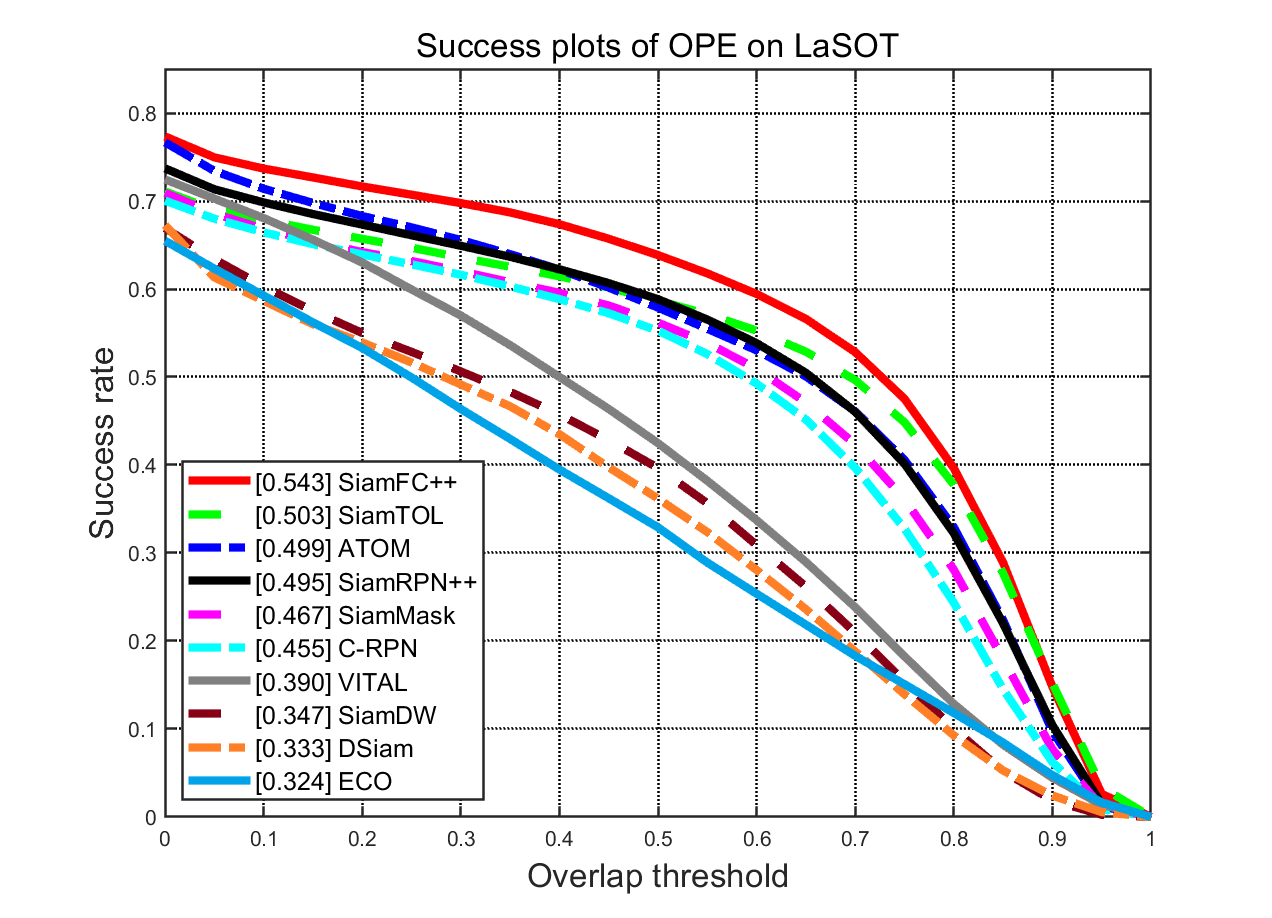}
		\includegraphics[width=0.49\linewidth]{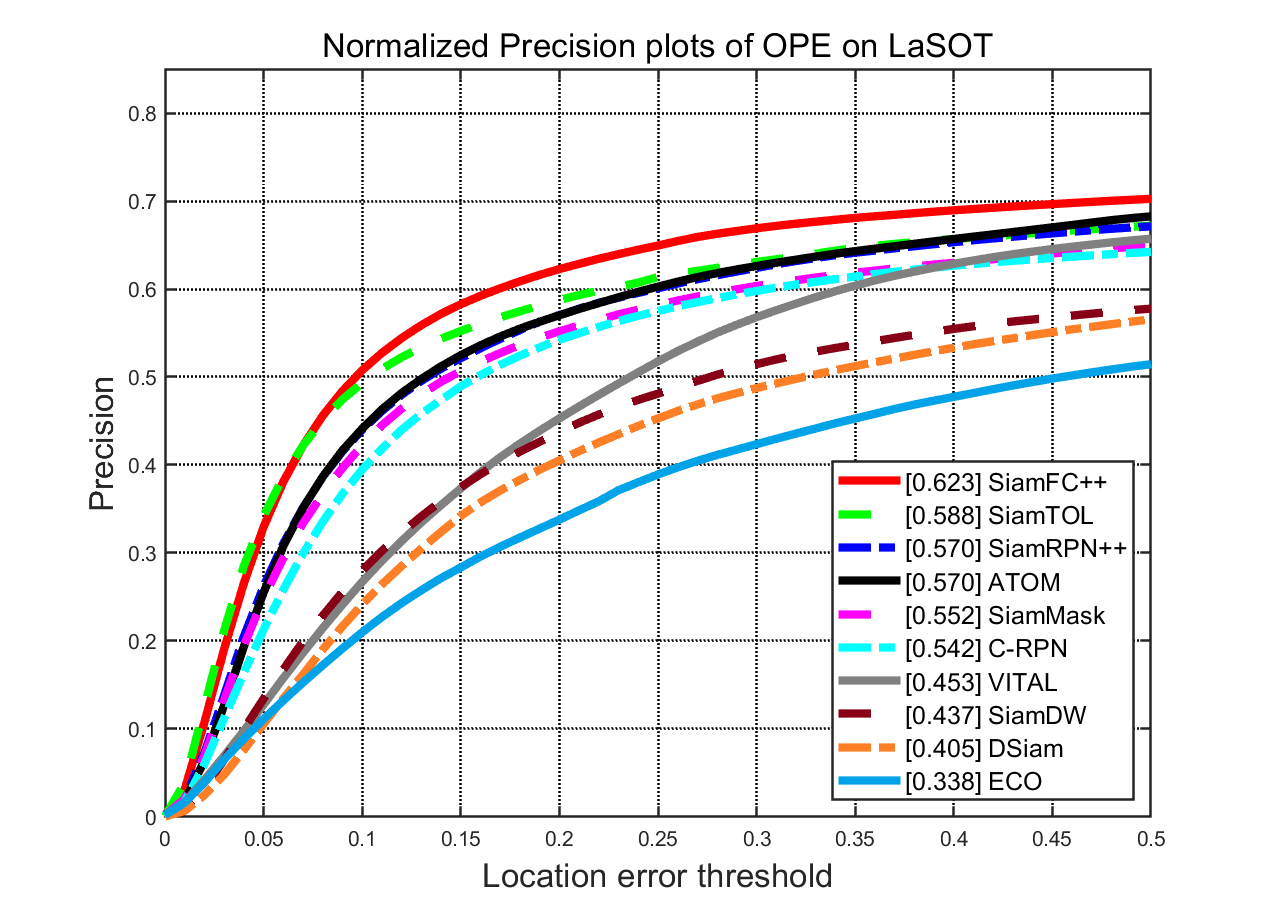}
	\end{center}
	\caption{Success and normalized precision plots on LaSOT dataset}
	\label{fig:auc_laSOT}
\end{figure}
\textbf{UAV123 Dataset \cite{Conference39}.} As a typical aerial tracking benchmark, the dataset contains 123 annotated sequences captured from Unmanned Aerial Vehicles. These sequences cover abundant challenging factors with an average length of 915 frames. Following the OTB2015, we adopt Precision score and AUC score to evaluate compared trackers, and the results are demonstrated on Figure \ref{fig:auc_UAV123}. Compared with other state-of-the-art methods, our tracker achieves pretty satisfactory performance.

\textbf{LaSOT Dataset \cite{Conference36}.}  We also validate the proposed approach on the LaSOT dataset, which is high-quality, large-scale dataset for single object tracking. There are 1400 sequences in total and 280 videos in the testing dataset with an average sequence length of more than 2,500 frames. The dataset is a great challenge to the short-time trackers. we display success and normalized precision plots in Figure \ref{fig:auc_laSOT}. The proposed updatable tracker ranks second in terms of both AUC and normalized precision scores, and outperforms SiamRPN++ tracker by $1.8\%$ on normalized precision.

\textbf{GOT-10k Dataset \cite{Journal7}.} This dataset is a large-scale high-diversity benchmark with more than 10k video sequences. Meanwhile, 180 annotated sequences are adopted as testing dataset, which consists of 84 kinds of objects in the wild with diverse motion patterns. We compare the performance of different trackers using the average overlap, and success rates corresponding to two overlap thresholds of 0.5 and 0.75. According to Table \ref{tab:got10k}, we find that our tracker outperforms most of previous methods with the highest speed. 

\subsection{Ablation Study.} 
\textbf{Online Update.} To explore the effect of online update, we compared the SiamTOL tracker with the baseline Siamese network without update. As shown in Table \ref{tab:abalation}, We found that our tracker still obtains 3.7$\%$ improvements on VOT2018 and 1.1$\%$ improvements on OTB100, even if trained only with a single overall loss. For the final version of our tracker, it outperforms the baseline Siamese tracker by 9.3$\%$ on VOT2018 and 2.4$\%$ on VOT100. These results depict that the presented two-stage one-shot learning is very effective in updating the Siamese tracker.

\textbf{Multi-aspect training losses.} We perform an ablation study on the training losses in diverse aspects to demonstrate the impact of multi-aspect loss training. It is easy to observe that the EAO on VOT2018 and the AUC on OTB100 respectively increases 2.8$\%$ and 0.7$\%$ via introducing the basic loss in addition to the overall loss, which indicates the basic loss is useful for improving training quality. Besides, adopting the update loss also is effective to promoting training performance, although its role is inferior to the basic loss since it violates the fact that the initial exemplar is the most important reference during tracking. After aggregating three losses in diverse aspects, the performance of tracker steadily increases, with improvements of 5.6$\%$ on VOT2018 and 1.3$\%$ on OTB100.

\textbf{Fusing network.} We perform a comparison between the standard direct-connected network and the presented residual network to study which is the more suitable fusion module.  According to Table \ref{tab:abalation}, the residual network gains 3.2$\%$ increasements on VOT2018 and 0.9$\%$ increasements on OTB100. This demonstrates that the best update strategy is to adjust the initial template with the latest object appearance, because it is the only reliable reference for tracking.

\begin{table}	
	\begin{center}
		\setlength{\tabcolsep}{1.0mm}{		
			\begin{tabular}{c|ccc|c|c|c}
				\toprule[1.5pt] 
				Updater & L-A & L-B & L-U &Fusion & VOT2018 &OTB100\\
				\midrule[1pt] 
				No &  &  &  &  & 0.366 & 0.672\\
				TOL & \checkmark &  &  & Res & 0.403 & 0.683 \\
				TOL & \checkmark & \checkmark &  & Res & 0.431 &0.689 \\
				TOL & \checkmark & & \checkmark & Res & 0.417 &0.684 \\
				\midrule[1pt] 
				TOL & \checkmark & \checkmark & \checkmark & Std & 0.427 & 0.687 \\
				\midrule[1pt] 
				TOL & \checkmark & \checkmark & \checkmark & Res & \textcolor[rgb]{1,0,0}{0.459} &\textcolor[rgb]{1,0,0}{0.696} \\
				\bottomrule[1.5pt]
		\end{tabular}}
	\end{center}
	\caption{Ablation study of the proposed tracker on VOT2018 and OTB2015. L-A, L-B, L-U represent the overall loss, basic loss and update loss described in sect. \ref{losses}, respectively. Fusion denotes the type of feature fusion network, where Std represents the direct-connection network without residual branch, and Res represent the residual scheme described in our paper.}	
	\label{tab:abalation}
\end{table}

\section{Conclusions}
In this paper, we exploit the potency of one-shot learning, and propose an effective two-stage one-shot learning scheme for online update. According to it, an updatable Siamese tracker named as SiamTOL is presented, which can complete reliable online update of its own. The object features in diverse stages are combined with the initial exemplar by residual learning to generate a more suitable template for current object. we also presented a multi-aspect loss function to describe how to efficiently train the model end-to-end. Extensive experiments on six tracking benchmarks demonstrate that the SiamTOL tracker can achieve state-of-the-art performance with a very tiny speed debasement, which confirms the effectiveness of two-stage one-shot learning on online update.

{\small
\bibliographystyle{ieee_fullname}
\bibliography{egbib}
}

\end{document}